\pdfoutput=1

\documentclass[11pt]{article}

\usepackage{EMNLP2022}

\usepackage{times}
\usepackage{latexsym}
\usepackage{xspace}
\usepackage{graphicx}
\usepackage{subcaption}
\usepackage{xcolor}
\usepackage{mathtools}
\usepackage{amssymb}%
\usepackage{pifont}%
\usepackage{soul}
\usepackage[T1]{fontenc}

\usepackage[utf8]{inputenc}

\usepackage{microtype}

\definecolor{atomictangerine}{rgb}{1.0, 0.6, 0.4}
\definecolor{babypink}{rgb}{0.96, 0.76, 0.76}
\definecolor{babyblueeyes}{rgb}{0.63, 0.79, 0.95}

\newenvironment{enumeratesquish}{\begin{list}{\addtocounter{enumi}{1}\labelenumi}{\setlength{\itemsep}{0em}\setlength{\labelwidth}{0.5em}\setlength{\leftmargin}{\labelwidth}\addtolength{\leftmargin}{\labelsep}}}{\end{list}\setcounter{enumi}{0}}

\newcommand{\dataset}{\textsc{TVStoryGen}\xspace}
\newcommand{\datasetsum}{\textsc{TVStorySum}\xspace}
\newcommand{\storium}{STORIUM\xspace}
\newcommand{\writing}{WritingPrompts\xspace}
\newcommand{\tvmegasite}{TVMegaSite\xspace}
\newcommand{\tms}{TMS\xspace}
\newcommand{\fandom}{Fandom\xspace}
\newcommand{\fd}{FD\xspace}

\newcommand{\kevin}[1]{\textcolor{blue}{\bf \small [ #1 --K]}}
\newcommand{\mingda}[1]{\textcolor{red}{\bf \small [ #1 --M]}}

\renewcommand{\kevin}[1]{}{}
\renewcommand{\mingda}[1]{}{}

\title{TVRecap: A Story Generation Dataset with Character Descriptions\\for Television Shows}
\title{TVStoryGen: A Dataset for Generating Stories\\with Character Descriptions}

\author{Mingda Chen\thanks{~~Work done while the author was at Toyota Technological Institute at Chicago.}\qquad Kevin Gimpel$^\dagger$\\
$^\dagger$Toyota Technological Institute at Chicago, Chicago, IL, 60637, USA\\
  {\tt \{mchen,kgimpel\}@ttic.edu}\\}

\begin{document}
\maketitle
\begin{abstract}
We introduce \dataset, a story generation dataset that requires generating detailed TV show episode recaps from a brief summary and a set of documents describing the characters involved. Unlike other story generation datasets, \dataset contains stories that are authored by professional screenwriters and that feature complex interactions among multiple characters. Generating stories in \dataset requires drawing relevant information from the lengthy provided documents about characters based on the brief summary.
In addition, we propose to train reverse models on our dataset for evaluating the faithfulness of generated stories.
We create \dataset from fan-contributed websites, which allows us to collect 26k episode recaps with 1868.7 tokens on average.
Empirically, we take a hierarchical story generation approach and
find that the neural model that uses oracle content selectors for character descriptions demonstrates the best performance on automatic metrics, showing the potential of our dataset to inspire future research on story generation with constraints. Qualitative analysis shows that the best-performing model sometimes generates content that is unfaithful to the short summaries, suggesting promising directions for future work.
\footnote{Data is available at \url{https://github.com/mingdachen/TVRecap}}

\end{abstract}

\section{Introduction}

Story generation is the task of generating a coherent narrative. Due to its open-ended nature, increasing efforts have been devoted to  constrained settings to facilitate reliable evaluation of computational models, such as generating stories from short prompts \cite{fan-etal-2018-hierarchical} and story continuations \cite{mostafazadeh-etal-2016-corpus} with various constraints \cite{akoury-etal-2020-storium}. In this work, we are interested in generating stories that accord with descriptions about the characters involved. The task is akin to writing stories based on true events or historical figures. For example, when writing historical fiction, writers use facts in biographies of historical figures (i.e., character descriptions) \cite{brown1998historical}. In a similar vein, cognitive psychologists observed that in order for narrative text to be compelling, it has to base its characters on real-world details such that readers can form emotional attachments to them even if the events occurring in the text are not realistic \cite{oatley1999fiction,green2003power}. In either case, computational models for this task can offer assistance in proposing possible stories constrained by relevant documents.

To this end, we create a story generation dataset \dataset that generates detailed TV show episode recaps from a brief summary of the episode and a set of lengthy character descriptions.
We construct \dataset from fan-contributed websites, which allows us to collect 26k episode recaps covering a variety of genres. An example from \dataset is shown in Figure~\ref{fig:dataset_example}. The dataset is challenging in that it requires drawing relevant information from the lengthy character description documents based on the brief summary. Since the detailed episode recaps are constrained by character descriptions, it also can evaluate neural models’ ability to maintain consistent traits or goals of particular characters during generation.

In addition, to evaluate the faithfulness of the generated stories to the brief summaries, we propose a metric that uses the perplexities from reverse models trained on our dataset. The reverse models are trained by considering generating the brief summary from the detailed recap. Human evaluation shows that this automatic metric outperforms competitive baselines in evaluating faithfulness.

Empirically, we characterize the dataset with several nearest neighbour methods and oracle models, finding that the use of the brief summaries and the character descriptions generally benefits model performance. We find that our non-oracle models are competitive compared to nearest neighbour models, suggesting promising future directions. We also benchmark several large pretrained models on the summarization version of our dataset, finding that they perform worse than an extractive oracle by a large margin despite the fact that the dataset favors abstractive approaches. Human evaluation reveals that without character descriptions, models tend to dwell on each event separately rather than advancing the plot, whereas using character descriptions improves the interestingness of the generated stories. Qualitatively, we show that models are able to generate stories that share similar topics with the summaries, but they may miss events in the summaries, leading to unfaithful generations.

We summarize our contributions below:
\begin{enumeratesquish}
\item We construct a story generation dataset of 26k instances and show (both qualitatively and quantitatively) that it %
has several unique challenges.

\item We show that the reverse models trained on our dataset can be used in evaluation for the original dataset, namely to determine whether generated stories are faithful to their input summaries.

\item We empirically characterize the story generation dataset and the summarization version of our dataset with several nearest neighbour methods, oracle models, and pretrained models, showing the challenges of these tasks and suggesting future research directions.

\end{enumeratesquish}
\section{The \dataset Dataset}

\begin{figure*}[t]
    \centering
    \includegraphics[scale=0.5]{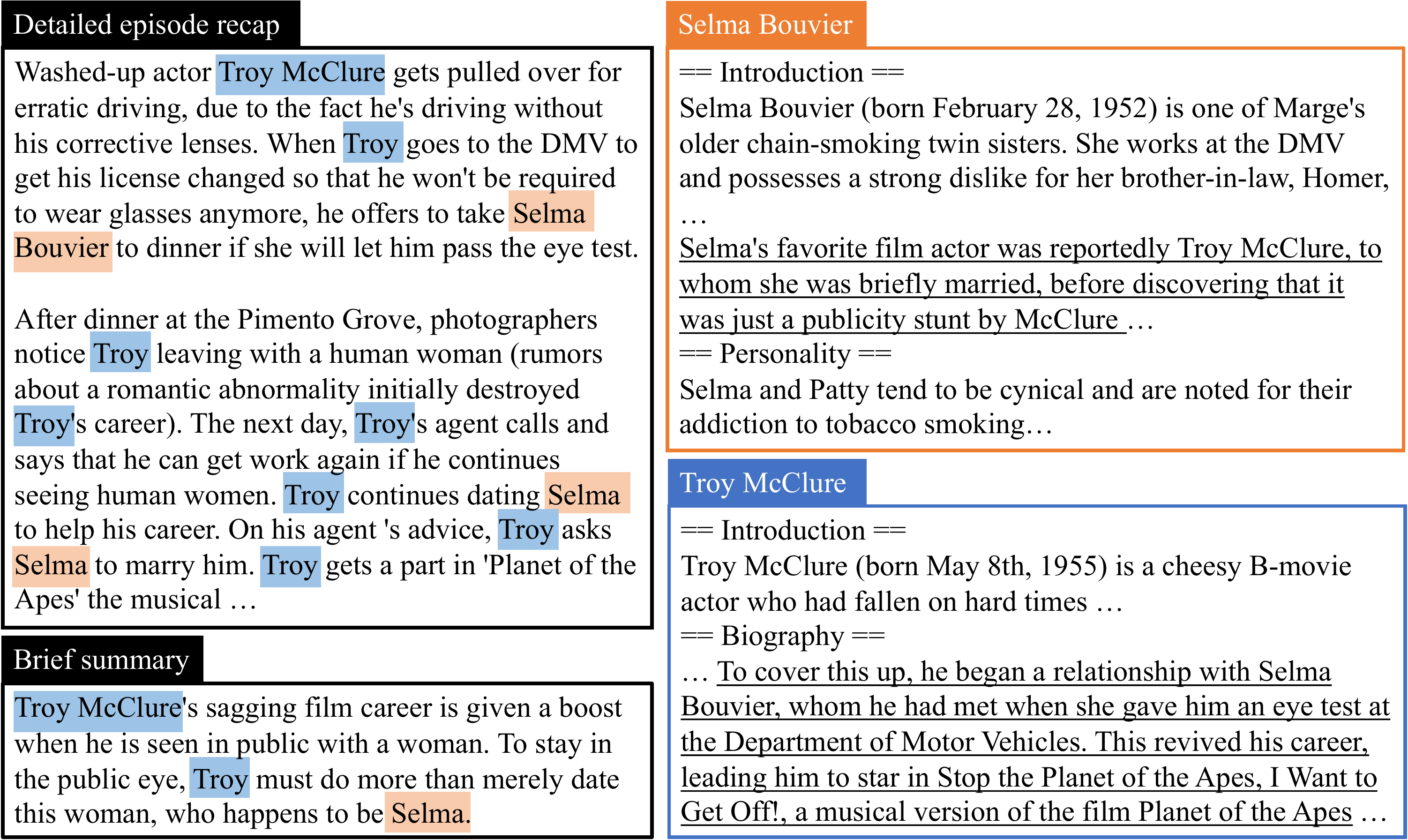}
    \caption{Excerpts from a \dataset instance that corresponds to the episode ``A Fish Called Selma'' in the TV show ``The Simpsons''. Colored texts are mentions of  characters. Texts surrounded by ``=='' in the character descriptions are section titles. We underline texts in the character descriptions that are closely related to the detailed recap. The shown part of the story includes complex interactions between %
    \colorbox{atomictangerine}{Selma\vphantom{y} Bouvier} and \colorbox{babyblueeyes}{Troy McClure}, which requires models to use relevant information from the lengthy character descriptions. There are six characters involved in this story but we only show parts of the detailed recap and character descriptions for brevity.}\vspace{-0.5em}
    \label{fig:dataset_example}
\end{figure*}

\begin{table*}[t]
    \centering
    \small
    \begin{tabular}{|c|r|r|r|r|r|c|}\hline
         & \# stories & \# tokens per story & \# char. per story & \# total unique char. & \# tokens per char. desc. \\\hline
        ROCStories & 98.2k & 88.0 & - & - & -   \\\hline
        \writing & 303.4k & 735.0 & - & - & -  \\\hline
        \storium & 5.7k & 19.3k & 4.6 & 26.2k & 269.0  \\\hline
        \dataset & 29.0k & 1868.7 & 16.7 & 34.3k & 1553.4  \\\hline
    \end{tabular}
    \caption{Statistics for story generation datasets with prompts as part of inputs. \dataset has moderate numbers of stories, moderate lengths of stories, long character descriptions, and a large number of total unique characters and characters per story.}
    \vspace{-1.5em}
    \label{tab:story_generation_dataset_compare}
\end{table*}

\begin{table}[t]
    \centering\small\setlength{\tabcolsep}{3pt}
\begin{tabular}{|l|r|r|}\hline
& \fd & \tms \\\hline
avg. \# tokens in summaries & 56.7 & 366.6 \\
avg. \# tokens in detailed recaps & 1291.7 & 2375.3 \\
avg. \# tokens in char. desc. per char. & 702.7 & 2300.9 \\
avg. \# tokens in char. desc. per ep. & 10891.9 & 40956.0  \\
avg. \# characters & 15.5 & 17.8 \\\hline
\end{tabular}
    \caption{Dataset statistics for \dataset. See the appendix for more details.}
    \label{tab:detailed_dataset_stats}
    \vspace{-1.5em}
\end{table}

In this section, we describe how we construct \dataset and compare it to other story generation datasets. 
An instance in \dataset is comprised of three components: (1) a detailed episode recap, (2) a brief summary of the episode, and (3) character descriptions, i.e., a set of documents describing the characters involved in the episode. The detailed episode recap delineates the events that occurred in the corresponding episode, which is usually written by fans after watching the episode. The documents about the characters contain biographical details and possibly personality traits. The summary either summarizes the whole episode or talks about the setup of the episode (to avoid spoilers). 

An example instance is shown in Figure~\ref{fig:dataset_example}, which comes from an episode of the TV show ``The Simpsons''. To demonstrate the level of detail of the character descriptions, we include a relatively complete character description for Selma Bouvier in the appendix, where the description consists of five sections: ``Introduction'', ``Personality'', ``Physical Appearance'', ``Relationships'', and ``Non-canon Appearance''. Each section's text is typically hundreds of word tokens long, covering many specifics.

As there are relevant details mentioned in the character descriptions, generating the detailed recap requires drawing information from the lengthy character descriptions. Moreover, due to the fact that the brief summary sometimes only depicts the setup of the episode, completing the story also necessitates using the information in the character descriptions. That is, the character description information is expected to be helpful for both filling in details that are not present in the brief summary as well as, for some of the instances, generating a plausible ending for the story. While the character descriptions could include some of the details in output stories, it is still challenging for models to retrieve salient information from the lengthy character descriptions and effectively integrate it into the discourse of the stories.

\subsection{Dataset Construction}
\label{sec:dataset_construction}
We construct \dataset from two fan-contributed websites: \fandom\footnote{\url{https://www.fandom.com/}} (\fd) and \tvmegasite\footnote{\url{http://tvmegasite.net/}} (\tms). We collect brief summaries and detailed episode recaps for several long-running soap operas from \tvmegasite and other TV shows from \fandom. We collect character descriptions from \fandom.\footnote{Data from Fandom is available under Creative Commons licenses and we have received permission from the owners of \texttt{tvmegasite.net} to publish their data for use with attribution.}  Since the pages on \fandom have hyperlinks pointing to the character pages, we use the hyperlinks to connect episodes to the characters involved. For \tvmegasite, where there are no such hyperlinks, we use string matching to find the characters.
To ensure the quality of this dataset, we filter out episodes based on several criteria. See the appendix for more details on the criteria and the string matching algorithm.

We report detailed statistics about \dataset in Table~\ref{tab:detailed_dataset_stats}. 
As shown in the table, there are systematic differences between \fd and \tms in terms of length of detailed episode recaps, summaries, and character descriptions, among others. We note that the character descriptions for \tms also come from \fd. Considering the differences, we train and evaluate models on the two splits separately in experiments. Since the summaries in \fandom are shorter and likely only depict the setups of the detailed recaps, we conduct a human evaluation to check the fraction of setups in the summaries, finding that 61.7\% of the summaries are setups.\footnote{We sample 60 episodes with 2 episodes per show.}

\begin{table}[t]
    \centering\small
\begin{tabular}{|l|ccc|}
\hline
\fandom & Train & Dev & Test \\ \hline
\# shows & 106 & 104 & 106 \\
\# episodes & 10833 & 1320 & 1430 \\
\hline
\tvmegasite & Train & Dev & Test \\ \hline
\# shows & 7 & 7 & 9 \\
\# episodes & 11586 & 1452 & 2392 \\
\hline
\end{tabular}
    \caption{Statistics of train/dev/test splits for \fandom and \tvmegasite.}
    \label{tab:split_stat}
    \vspace{-1.5em}
\end{table}

We verify the diversity of topics covered in \dataset, finding that \fd covers far more genres than \tms with the most frequent occupying only 15\% of episodes. See the appendix for more details. We randomly split the datasets into train/dev/test sets. For \tms, we additionally filter out instances if the overlap ratio of TV show characters appearing in the summary and the detailed recap is lower than 85\%. This extra filtering step ensures alignment between the summaries and detailed recaps. The final statistics of the splits is shown in Table~\ref{tab:split_stat}.

\subsection{Dataset Comparison}

We compare \dataset to other story generation datasets in Table~\ref{tab:story_generation_dataset_compare}. Unlike ROCStories \cite{mostafazadeh-etal-2016-corpus} and \writing \cite{fan-etal-2018-hierarchical} where the inputs to models are either the first few sentences or short prompts, \dataset has character descriptions as extra constraints, making the task of generating the reference stories from the inputs less open-ended and therefore more feasible.

Since \storium \cite{akoury-etal-2020-storium} has character descriptions and other information as constraints, it is the most comparable resource to \dataset. Our dataset differs from \storium in that \dataset has more stories, more characters, and longer, more diverse, detailed character descriptions. See the appendix for more detailed discussion.

Moreover, models trained on \dataset can potentially complement those from \storium by merging all the characters' turns in \storium into a coherent narrative. We also quantitatively demonstrate that the source inputs and the output stories are more directly related in our dataset than \storium. Details are in the appendix.

\subsection{Dataset Challenges}

\begin{table}[t]
\setlength{\tabcolsep}{4pt}
    \centering\small
    \begin{tabular}{|p{0.45\textwidth}|}\hline
         ...Meanwhile, Patty and {\bf Selma} have received a promotion at the DMV and have more disposable income. As a last resort, {\bf Homer} asks the two if they will lend him the money. They agree, but he must become their loyal servant. Patty and {\bf Selma} make {\bf Homer}'s life a living hell...\\\hline\hline
        ...The next day, news of {\bf Homer}'s ``death'' spreads across Springfield, and Marge starts getting condolences from prominent Springfieldians. Patty and {\bf Selma} offer their condolences in the form of a tombstone celebrating {\bf Homer}’s death...\\\hline
    \end{tabular}
    \caption{Two excerpts in detailed recaps from \dataset that correspond to different episodes in the TV show ``The Simpsons''. The excerpts involve interactions between Homer and Selma where Selma consistently shows a strong dislike for Homer, matching the character description in Figure~\ref{fig:dataset_example}. }
    \vspace{-0.5em}
    \label{tab:challenge_example}
\end{table}

\dataset poses several challenges for story generation models. The first challenge stems from the long lengths of the inputs and outputs. Specifically, the average instance in \dataset has a story of 1.8k tokens and character descriptions of more than 10k tokens (see Table~\ref{tab:detailed_dataset_stats}). In contrast, story generation and multi-document summarization datasets may have lengthy inputs or outputs, but rarely both (see Table~\ref{tab:story_generation_dataset_compare} and appendix~\ref{sec:compare_to_other_summ_datasets} for detailed statistics). The long inputs and outputs make it challenging to design models that can effectively integrate lengthy character descriptions into the generation process of long and coherent stories.

The other set of challenges relates to consistency in character modeling. Since the episode recaps are constrained by character descriptions, the dataset provides opportunities to evaluate neural models' ability to maintain consistent personalities or goals of particular characters during generation. The consistency of personalities and goals is related to the notion of ``character believability'' \cite{bates1994role,riedl2010narrative}, which has been deemed important for composing convincing stories. We illustrate %
this challenge with two excerpts in Table~\ref{tab:challenge_example}: the strong dislike that Selma has shown for Homer matches her description and is consistent across episodes. Solving this challenge requires models to first identify related information in the lengthy character descriptions based on the plot and integrate it into the generated narrative. We aim to incorporate this idea into the design of our models.

\section{Approach}

We follow \citet{fan-etal-2019-strategies} to take a hierarchical story generation approach. The generation process is broken into two steps that use two separately parameterized models: a text-to-plot model and a plot-to-text model. The text-to-plot model first generates detailed plots based on the inputs, and then conditioned on the plots the plot-to-text model generates detailed stories. In this paper, we define the plots as linearized semantic role labeling (SRL) structures. More details on SRL are in the appendix.
For example, a plot may be as follows:
\begin{align*}
    &\langle\text{VERB}\rangle\text{spots}\langle\text{ARG0}\rangle\text{Mummy Pi}\\ &\langle\text{ARG1}\rangle\text{how messy the car is}\langle\text{SEP}\rangle\\ &\langle\text{VERB}\rangle\text{clean}\langle\text{ARG0}\rangle\text{they}
    \langle\text{ARG1}\rangle\text{the car}
\end{align*}
where $\langle\text{SEP}\rangle$ is a special token used to separate SRL structures for different sentences.

\paragraph{Text-to-Plot Model.} During training, we use the oracle plots, i.e., the SRL tags extracted from the reference recaps. During test time, we use BM25 \cite{robertson1995okapi} to find the most similar plot in the training set from the same show based on either the summaries or the detailed recaps (as an oracle baseline).\footnote{We find the plots generated by neural models to be of lower quality.} If a show is not present in the training set, we search over the whole training set.

\paragraph{Plot-to-Text Model.} Our models are based on the sequence-to-sequence transformer architecture \cite{attenion_is_all_you_need}. 
Similar to \citet{rothe-etal-2020-leveraging} that uses pretrained BERT-like models to initialize sequence-to-sequence models, we use the pretrained RoBERTa-base model \cite{liu2019roberta} as the decoder.\footnote{We chose RoBERTa over GPT-2 \cite{radford2019language} because BERT-style models outperform GPT-2 in the encoder-decoder setting in the results reported by  \citet{liu2019roberta}.} For the encoder, we choose to use a one-layer randomly initialized Longformer \cite{beltagy2020longformer} due to the lengthy inputs and computational constraints. We randomly initialize other parameters and finetune the whole model during training. 

Given a plot, we use the neural models to generate sentence by sentence as we find this yields better performance than generating the whole detailed recap. When doing so, we concatenate the SRL tags for the adjacent sentence of the target sentence with the SRL tags for the target sentence. This gives similar performance to showing the SRL tags for the entire detailed recap but is much more efficient (due to shorter sequence lengths). %
Because the character descriptions are lengthy, we use BM25 to retrieve the most salient information from character descriptions (i.e., one sentence) for each sentence in the detailed recap. We note that during test time, when the text-to-plot model retrieves plots from the training set, we also use the corresponding selected character descriptions.

The pipeline that retrieves relevant character information and then adapts it based on the plot is the first step that we take to simulate a writing system that can dynamically update its belief about particular characters based on the given relevant documents. This differs from prior work on entity representations for story generation \cite{clark-etal-2018-neural} that does not consider character descriptions as we do. 

The inputs to plot-to-text models contain two sources: plots and character descriptions. Since there could be multiple entries corresponding to different characters in the character descriptions, we include a type embedding to differentiate different entries and sources in the input. Similar approaches have been used to represent table entries in neural models \cite{dhingra-etal-2019-handling}. For example, for Figure~\ref{fig:dataset_example} the inputs are
\begin{align*}
    \langle\text{SEP}\rangle_0\text{Troy McClure's ...}_0\langle\text{SEP}\rangle_1\text{Selma Bouvier}\\
    \langle\text{SEP}\rangle_1\text{Selma's favorite film ...}_1\langle\text{SEP}\rangle_2\text{...}
\end{align*}
where the subscripts indicate the ID of the type embedding and we always prepend the character names to the corresponding selected character descriptions. %
The final vector representation of the input is the summation of subword unit embeddings, positional embeddings, and the type embeddings. Conditioned on the input representations, we train the RoBERTa decoder on the reference recaps using a cross-entropy loss.

Due to computational constraints, for the Longformer encoder, we use the global attention on the $\langle\text{SEP}\rangle$ tokens, and use the encoded representations for the summary, the SRL tags, and the $\langle\text{SEP}\rangle$ tokens in character descriptions as the input to decoders.

\section{Experiments}

\subsection{Setup}

For evaluation, we report BLEU (BL) \cite{papineni-etal-2002-bleu}, ROUGE-1 (R1), ROUGE-2 (R2), and ROUGE-L (RL) \cite{lin-2004-rouge} scores. We additionally report perplexities of the summaries given the generated stories using a reverse model. We will refer to this metric as ``PL''. This metric evaluates the faithfulness of the generated stories to the summaries. Lower PL suggests better faithfulness. When computing PL, we use the Pegasus model \cite{pegasus-zhang20ae} finetuned on our dataset as it has the best test set perplexities. Details on model selection, decoding and hyperparameters are in the appendix.

When training the reverse model on our dataset, we simply use the detailed episode recap as the source input and the summary as the target output. In the appendix, we benchmark competitive pretrained models on the ``inverted'' dataset and compare the dataset to other summarization datasets, finding that it is a potentially valuable contribution for the summarization community.

\subsection{Results}

\begin{table*}[t]\small
    \centering\setlength{\tabcolsep}{6pt}
    \begin{tabular}{|l|r|r|r|r|r|}\hline
         & BL ($\uparrow$) & R1 ($\uparrow$) & R2 ($\uparrow$) & RL ($\uparrow$) & PL ($\downarrow$)  \\\hline
         \multicolumn{6}{|c|}{Development set results} \\\hline
        (NN) Nearest neighbour plot + summary + char. desc. & 7.1 & 40.7 & 11.0 & 39.6 & 32.3 \\
        (NN) Oracle plot & 21.2 & 52.8 & 24.1 & 51.8 & 23.1 \\
        (NN) Oracle plot + summary & 24.5 & 54.3 & 25.6 & 55.2 &  20.8 \\
        (NN) Oracle plot + summary + oracle char. desc. & \bf 28.4 & \bf 63.0 & \bf 32.8 & \bf 61.2 & \bf 17.9 \\
        \hline
         \multicolumn{6}{|c|}{Test set results} \\\hline
        (Return) reference & 100.0 & 100.0 & 100.0 & 100.0 & 12.9 \\
        (Return) oracle plot  & 3.6 & 43.9 & 19.7 & 41.9 & - \\
        (Return) oracle plot + summary & 5.4 & 48.5 & 20.5 & 46.2 &  - \\
        (Return) oracle plot + oracle char. desc. & 1.2 & 11.0 & 4.6 & 10.6 & - \\
        (Return) oracle plot + oracle char. desc. + summary & 1.2 & 11.0 & 4.7 & 10.6 & - \\\hline
        (Return) Nearest neighbour detailed recap & 5.1 & 41.1 & 9.3 & 39.6 & 31.2 \\
        (Return) Oracle nearest neighbour detailed recap & 4.8 & 41.2 & 10.8 & 39.9 & 28.5 \\\hline
        (NN) Nearest neighbour plot + summary + char. desc. & 6.0 & 41.7  & 10.7 & 40.3 & 28.0 \\
        (NN) Oracle plot + summary + oracle char. desc. & \bf 28.4  & \bf 63.2 & \bf 32.9 & \bf 61.5 & \bf 18.2  \\\hline
    \end{tabular}
    \caption{Results for \fandom in \dataset. The results for the return-input baselines and the neural models are indicated by``(Return)'' and  ``(NN)'' respectively. The best result in each column for each split (excluding the references) is boldfaced.}
    \label{tab:fandom_main_result}
\end{table*}

\begin{table*}[t]\small
    \centering\setlength{\tabcolsep}{6pt}
    \begin{tabular}{|l|r|r|r|r|r|}\hline
         & BL ($\uparrow$) & R1 ($\uparrow$) & R2 ($\uparrow$) & RL ($\uparrow$) & PL ($\downarrow$)  \\\hline
         \multicolumn{6}{|c|}{Development set results} \\\hline
        (NN) Nearest neighbour plot + summary + char. desc. & 10.7 & 43.5 & 14.9 & 42.9 & 20.7 \\
        (NN) Oracle plot & 26.4 & 60.5 & 34.0 & 60.0  & 17.0 \\
        (NN) Oracle plot + summary & 28.3 & 64.3 & 36.1 & 63.9 & 16.4  \\
        (NN) Oracle plot + summary + oracle char. desc. & \bf 30.9 & \bf 68.3 & \bf 44.0 & \bf 67.5 & \bf 15.7 \\
        \hline
         \multicolumn{6}{|c|}{Test set results} \\\hline
        (Return) Reference & 100.0 & 100.0 & 100.0 & 100.0 & 13.9 \\
        (Return) Oracle plot  & 7.1 & 53.1 & 22.3  & 52.4  & - \\
        (Return) Oracle plot + summary & 13.6 & 62.9 & 25.6 & 62.0 &  - \\
        (Return) Oracle plot + oracle char. desc. & 1.1 & 12.3 & 5.1  & 11.9 & - \\
        (Return) Oracle plot + oracle char. desc. + summary & 1.2 & 12.4 & 5.1 & 12.0 & - \\\hline
        (Return) Nearest neighbour detailed recap & 6.6 & 49.8 & 16.0 & 49.2 & 26.3 \\
        (Return) Oracle nearest neighbour detailed recap & 7.5 & 49.9 & 18.5 & 49.3 & 25.8 \\\hline
        (NN) Nearest neighbour plot + summary + char. desc. & 7.3 & 50.6 & 17.6 & 49.8 & 25.3 \\
        (NN) Oracle plot + summary + oracle char. desc. &  \bf 28.1 & \bf 67.0 &\bf 40.9  & \bf 66.2 & \bf 18.3  \\\hline
    \end{tabular}
    \caption{Results for \tvmegasite in \dataset. The results for the return-input baselines and the neural models are indicated by ``(Return)'' and ``(NN)'' respectively. The best result in each column for each split (excluding the references) is boldfaced.}
    \vspace{-1.5em}
    \label{tab:tvmegasite_main_result}
\end{table*}

We report results for \fd and \tms in Tables~\ref{tab:fandom_main_result} and \ref{tab:tvmegasite_main_result},  respectively. We report several return-input baselines on the test sets to show the benefits of using neural models as plot-to-text models. We report PL on the test sets as an approximated lower bound of this metric. We do not report PL on return-input baselines as the output detailed recaps involve SRL sequences, which are not natural language, and therefore the results are not comparable to others.

On the development sets, adding summaries and oracle character descriptions generally improves performance by a significant margin, showing that the extra information aids generation.

Regarding the test set results, we find that (1) the return-input baselines show that the performances of our neural models are non-trivial; (2) while the oracle nearest neighbour baselines achieve competitive performance to our non-oracle neural models, the non-oracle neural models are consistently better than the non-oracle baselines, showing promising results for future research on our datasets. We note that the return-input baselines involving character descriptions display much worse results than other return-input baselines because they are lengthy, which leads to low precision.
\kevin{It's a little hard to connect this discussion with the results because it's so dense. I think it may help to use different naming in Tables 4-5 and 5.2 to indicate in the name which methods use neural models. If someone is looking through quickly and sees both ``Oracle plot'' and ``Return oracle plot'' they will not think that ``Oracle plot'' is actually a neural model. Also, maybe change ``Reference'' to ``Return reference''?}\mingda{updated}

\section{Analysis}

\subsection{Human Evaluation}

\begin{table}[t]
    \centering\small
    \begin{tabular}{|l|r|r|}\hline
        & \multicolumn{1}{|c|}{Relevancy} & \multicolumn{1}{|c|}{Interesting} \\\hline
        \multicolumn{3}{|c|}{Expert annotations}\\\hline
        Prefer summary & 60.0\% (30/50) & 40.0\% (20/50) \\
        Prefer char. desc. & 54.0\% (27/50) & 70.0\% (35/50) \\\hline
        \multicolumn{3}{|c|}{Crowdsourced annotations}\\\hline
        Prefer summary & 50.0\% (11/20) &  55.0\% (10/20)\\
        Prefer char. desc. & 55.0\% (11/20) & 55.0\% (11/20)\\\hline
    \end{tabular}
    \caption{Human annotation results analyzing the effect of including different components in the inputs. The percentage is the fraction of annotations that favor the models to include the corresponding component. The numbers in parentheses are the number of positive annotations divided by the total number of annotations.}
    \label{tab:human_annoation_result}
    \vspace{-0.5em}
\end{table}

\paragraph{Effect of Different Components in Our Dataset.} To measure the impact of including different components in \dataset, we conduct a human evaluation. We show two generated stories from different models along with the corresponding brief summary and ask annotators to choose which story they prefer according to two aspects: (1) which generation is more relevant to the summary; (2) which story is more interesting.

We make two comparisons: ``oracle plot'' vs. ``oracle plot+summary'' for studying the benefits of using summaries (``Prefer summary''), and ``oracle plot+summary'' vs. ``oracle plot+summary+oracle char.~desc.'' for studying the benefits of using character descriptions (``Prefer char.~desc.''). We sample instances from the \fd development set because the average lengths in \fd are shorter, and we only show annotators the first 100 tokens of the texts as we expect it to be challenging to annotate lengthy texts. %
We use Amazon Mechanical Turk (AMT) and collect 20 annotations per question for each comparison with 6 workers involved (shown in Table~\ref{tab:human_annoation_result} as ``crowdsourced annotations'').\footnote{To ensure annotation quality, we hire workers with master qualification and pay them with a target hourly wage of \$12. Details on annotation instructions are in the appendix.} We (the authors) also annotate 50 instances per comparison using the same interface as AMT (shown in the table as ``expert annotations''). While the crowdsourced annotations do not suggest clear benefits of using summaries and character descriptions, the expert annotations show that including the summary helps to improve relevancy but hurts the interestingness of the stories, whereas including character descriptions improves the interestingness despite the marginal benefits of improving the relevancy.
Recent work \citep{karpinska-etal-2021-perils} also found that compared to experts, AMT workers produce lower quality annotations for tasks like story generation.

When examining annotations, we find that models without using character descriptions tend to generate sentences that use the word ``but'' to negate what has been said in the earlier part of the sentence, leaving the sentence dwelling on each event separately rather than advancing the plot (see Sec.~\ref{sec:gen_examples} for examples).

\begin{table}[t]
    \centering\small
    \begin{tabular}{|l|r|}\hline
         &  \multicolumn{1}{|c|}{Acc.} \\\hline
        BLEU & 50.0 \\
        BLEURT & 54.5 \\
        PL (our metric) & \bf 61.4 \\\hline
    \end{tabular}
    \caption{Accuracies when evaluating the automatic metrics against human annotations. When computing the BLEU and BLEURT scores, we compare the generation against the brief summaries. The best performance in each column is in bold.}
    \label{tab:eval_pl_metric}
\end{table}

\paragraph{The PL Metric.} To verify the efficacy of our proposed metric PL, we compute accuracies between the PL metric and the human annotations (we use the expert relevancy annotations in human evaluation results). 
We consider BLEU and BLEURT \citep{sellam-etal-2020-bleurt} as baseline metrics by computing the generation against the brief summaries. When reporting results, we use the truncated generation to ensure consistency with human annotators. We show the results in Table~\ref{tab:eval_pl_metric}. We find that PL outperforms BLEU and BLEURT significantly, showing that PL is a promising metric for evaluating the faithfulness of generated story.

\subsection{Generation Examples}
\label{sec:gen_examples}
\begin{table*}[t]
\setlength{\tabcolsep}{5pt}
    \centering
    \small
\begin{tabular}{|p{0.22\textwidth}|p{0.21\textwidth}|p{0.22\textwidth}|p{0.25\textwidth}|}\hline
\multicolumn{1}{|c|}{Input summary} & \multicolumn{1}{|c|}{Reference} & \multicolumn{1}{|c|}{Oracle plot+summary} & \multicolumn{1}{{c}|}{Oracle plot+summ.+oracle char.} \\\hline
\ul{Elfman and Evergreen , after much struggle , lose to Rustyrose and his imagination - based Magic .} In the meantime , Gray , Lucy , Cana and Loke have been overpowered by Caprico alone . Loke decides to take him on by himself because of his mysterious Magic .
& \ul{Elfman and Evergreen encounter a cliff as they run away from Rustyrose 's Belcusas the Thunderclap .} Rustyrose appears shortly afterwards and expands on the idea of " The Ultimate World of Magic " , saying that all those who can not use Magic and the trash in the guilds are useless ...
& Elf episode begins with Elfman and Evergreen encounter a cliff in the middle of the forest , and the two begin to fight . All , all those those are going to use Magic to defeat them , and they will be able to defeat all of them . Ruth , however , has n't been able to outsmart them , and the two of them begin to fight ... 
& \ul{Elfman and Evergreen encounter a cliff as they run away from the danger of the Rustyrose 's attacks .} Evergreen explains that " all those who use Magic " will use the power of the Magic of the Seven . However , Rustyrose outsmarts them , stating that their Magic is useless against them , as they have already been defeated . He then finishes the two with Tower of Dingling .. \\\hline
\ul{Chuck is preparing for his new club opening and enlists Serena 's help} , but Blair begins to feel left out . Jenny , the new Queen at Constance , struggles between proving herself and her friendship with Eric , and Dan feels inferior after watching one of Olivia 's movies . Meanwhile , Lily tries to respect Rufus ' Halloween traditions .
& ... Blair is confident her plan will work , until \ul{Chuck calls and asks for Serena 's help in conducting the club opening .} He tells her he wants to open the next day on Halloween and that he does n't want Blair anywhere near the planning ...
& ... Chuck tells Blair that he 's not going to let her go , {\bf but that she 's going to be there for him} . He he opens the club and tells her that he 's going to see Serena at the loft , where she 's staying . She suggests an 80-year party party , {\bf but she says that 's not what she 's looking for} ...
& ... Blair goes to see Serena at the gallery , and she explains to her that she went to see Chuck . She tells her that she hired a party planner to help her hang out with Chuck , and Serena offers to help . \ul{At the VDW 's , Serena is conducting the club opening and Blair asks her to help .} He tells her that he does n't want Blair anywhere near the planning , and he wants her to be at the party for the night ...\\\hline
\end{tabular}
\caption{Excerpts from generation examples, which come from the TV shows ``Fairy Tail'' and ``Gossip Girl'' respectively. The highlighted texts are meaningless negations. We underline texts that describe similar events. The complete generations are in the appendix.}
\vspace{-0.5em}
\label{tab:gen_examples}
\end{table*}

We display the generation examples in Table~\ref{tab:gen_examples} where we find that generations from both models generally share similar topics and character names with the summaries and the references. For example, for the first instance, both generations are about a battle that concerns Elfman, Evergreen, and Rustyrose. However, as observed in the human evaluations, the ``oracle plot+summary'' model suffers from meaningless negation. For example, see the second generation example, where the highlighted texts keep negating the earlier plot development. While the ``Oracle plot+summ.+oracle char.'' model does not have this problem, it is still not faithful to the summary. Specifically, both the summary and the reference mention that Chuck needs Serena's help for his new club opening, but the generation states that ``Serena is conducting the club opening'' and ``Blair asks her to help''. This is likely caused by the model's inability to understand the states of each character (possibly due to the fact that our models generate at the sentence level) and to  effectively integrate multiple sources of information into a coherent narrative.
\section{Related Work}

Early methods in computational modeling for story generation rely on handwritten rules \cite{meehan1977tale,liu2002makebelieve} to structure narrative. Recent work has explored different approaches to improve the quality of story generation systems, including commonsense knowledge \cite{mao-etal-2019-improving,guan-etal-2020-knowledge}, automatically extracted key words \cite{peng-etal-2018-towards} and key phrases \cite{orbach-goldberg-2020-facts2story,rashkin-etal-2020-plotmachines}, event-based representations \cite{martin2018event}, and plot graphs \cite{li2013story}.

As our model involves plot generation and character modeling, it is related to work on plot planning \cite{riedl2010narrative,li2013story,martin2018event,yao2019plan,jhamtani-berg-kirkpatrick-2020-narrative}, character modeling \cite{clark-etal-2018-neural,liu2020character}, and the interplay between the two \cite{riedl2010narrative}. Our work is different in that it explicitly requires performing inference on lengthy documents about characters.

There have been other datasets built from TV shows, such as summarizing TV shows into character descriptions \cite{shi-etal-2021-descgen}, constructing knowledge bases \cite{chu2021knowfi}, summarizing TV show screenplays \cite{chen2021summscreen} and other textual data \citep{yu-etal-2016-unsupervised}, entity tracking \cite{chen-choi-2016-character,choi-chen-2018-semeval}, entity linking \citep{logeswaran-etal-2019-zero}, coreference resolution \cite{chen-etal-2017-robust,zhou-choi-2018-exist}, question answering \cite{ma-etal-2018-challenging,yang-choi-2019-friendsqa}, speaker identification \citep{ma-etal-2017-text}, sarcasm detection \citep{joshi-etal-2016-harnessing}, emotion detection \citep{zahiri2017emotion,hsu-ku-2018-socialnlp}, and character relation extraction \cite{yu-etal-2020-dialogue}.

\section{Conclusion}
We constructed a story generation dataset of 26k stories where each instance consists of a detailed episode recap, a summary, and several character descriptions. We quantitatively and qualitatively illustrate several unique challenging aspects of this dataset. In addition, we show that the reverse model trained on our dataset, which generates the summary from the detailed recap, is a competitive automatic metric for evaluating faithfulness of stories. We also propose a metric based on the summarization model trained on our dataset for evaluating the faithfulness of the generated stories to the summaries. Empirically, we benchmark several nearest neighbour models and oracle models, showing that the summaries and the character descriptions are helpful in generating better stories, which are verified by human evaluations.

\section{Limitations}

We highlight three limitations. Firstly, \dataset was derived from Fandom and \texttt{tvmegasite.net}. These two websites were contributed by fans to document details for popular TV shows. Due to the inconsistencies between our goal (i.e., story generation) and the goal of these websites, there could be considerable variations across instances regarding how well the data matches our purpose. For example, in Sec.~\ref{sec:dataset_construction}, we discuss the issue that some summaries do not include story endings.

The second is that our proposed models are relatively small-scale, with components parameterized by bag-of-words systems (i.e., BM25) due to computational constraints. Future work may explore using more powerful models, e.g., pretrained retrieval models like REALM \citep{pmlr-v119-guu20a} and RETRO \citep{borgeaud2021improving}.

The third limitation is human evaluation. In particular, we only have a small number of expert annotations. Future work may explore recruiting more experts from platforms like Upwork.\footnote{\url{https://www.upwork.com/}} Also, since we only show the beginning of a story to human annotators, it ignores the quality of story endings. Future work may explore a more robust human evaluation process, e.g., splitting stories into multiple parts and having several annotators evaluate different parts of stories.

\section*{Acknowledgements}
We would like to thank David Bamman and Matthew Sims for helpful discussions.

\clearpage
\bibliography{anthology,custom}
\bibliographystyle{acl_natbib}

\clearpage
\appendix

\section{Appendix}

\subsection{Example of Character Descriptions}
\begin{table*}[t]
    \centering\small
    \begin{tabular}{|l|p{0.8\textwidth}|}\hline
        \textbf{Section} & \textbf{Description} \\\hline
        Introduction & Selma Bouvier-Terwilliger-Hutz-McClure-Discothèque-Simpson-D'Amico (née Bouvier) is one of Marge's older chain-smoking twin sisters. She works at the DMV and possesses a strong dislike for her brother-in-law, Homer, although on extremely rare occasions she shows kindness towards him and seems to tolerate him. She seems to despise Homer slightly less than her twin sister, Patty Bouvier.

Selma Bouvier was born two minutes before Patty. Due to a childhood bottle rocket accident, she lost her sense of taste and smell. Selma is 43 years old and is four years older than Marge ... \bf (859 tokens)\\\hline
Personality & Selma and Patty tend to be cynical and are noted for their addiction to tobacco smoking. In one episode during Season 26, they temporarily live with Homer. Homer places a ban on smoking, so instead, they smoke electric cigarettes, which they hate. Selma has presumably switched to chewing tobacco after adopting Ling, although the episode “Puffless” disproves this. They have a strong, mutual (and reciprocated) dislike for their brother-in-law. Selma and Patty are shown to be older than Homer and Marge, but a birth date has not been given. It is presumed they are in their mid to late 40s since Selma has gone through menopause and the twins are shown as preteens or teenagers when Marge is around Lisa’s age. She enjoys getting foot massages as she is shown getting them constantly. She weighs 168 pounds exact and is considered overweight like Patty, though they are not as fat as Homer. As teenagers and children, they are average weight, while Homer was still fat... \bf (917 tokens) \\\hline
Physical Appearance &  She is tall and overweight with a similar body type to Patty. She wears a blue sleeveless dress and round earrings. Her hair is long and curly, worn in an m-shape (although she wore different hairstyles when she was younger, it’s always been longer than Patty’s). She is actually a blonde, in contrast to Patty‘s red hair, meaning that they are in fact fraternal twins. The gray coloring is from cigarette ashes. On special occasions, Selma will wear earrings in the form of the letter "S", further distinguishing her from her twin, who wears triangle earrings ... \bf (136 tokens) \\\hline
Relationships & Despite being twins, Patty and Selma have very different track records when it comes to finding dates. According to Marge, Patty chose a life of celibacy (possibly because she’s a lesbian), while Selma had celibacy thrust upon her. Selma is a heterosexual. Her standards are extremely low, as evidenced by her comments on Mr. Burns: "Single, eh? Well, he passes the Selma Test,” despite him being old enough to be her great-grandfather. Oddly, she was grossed out when Hans Moleman tried to kiss her ...  \bf (676 tokens) \\\hline
Non-canon Appearance & At age 55, Selma threatens to stuff Edna Krabappel's hat down her throat if she catches the bouquet at "Lisa's Wedding". This episode is not canon since it aired before Edna’s death and in real time, Edna would not have been at the wedding. In "Holidays of Future Passed", at age 72, she has a love-bot, who runs away with Patty's love bot, much to their annoyance ... \bf (697 tokens) \\\hline

    \end{tabular}
    \caption{Character Descriptions for Selma Bouvier from the TV show ``The Simpsons''. We omit part of the text in each section for brevity. The numbers at the end of each section text are the number of omitted word tokens.}
    \label{tab:example_char_desc}
\end{table*}

In Table~\ref{tab:example_char_desc}, we show detailed character descriptions for Selma Bouvier from the TV show ``The Simpsons''.

\subsection{Dataset Construction}

\paragraph{String Matching Algorithm.} For example, for the character name ``John Doe'', valid mentions are itself, ``John'', ``J.D.'', and ``JD'' due to the writing style on \tvmegasite. While this matching algorithm may lead to extra characters aligned to particular episodes, it at least includes all characters that are actually involved in the episode.

\paragraph{Episode Filtering Criteria.} We filter out episodes if (1) an episode contains fewer than 3 characters (to avoid stories that do not involve many character interactions); or (2) the detailed recap has fewer than 200 word tokens (ensuring that stories have enough details); or (3) the brief summary has fewer than 20 word tokens (to ensure that there is sufficient information given as the input).

\subsection{Detailed Statistics for \dataset}

\paragraph{Detailed Statistics for \fd and \tms.}

\begin{table}[t]
    \centering\small
\begin{tabular}{|l|r|r|}\hline
& \fd & \tms \\\hline
number of shows & 106 & 9 \\
number of episodes & 13583 & 15430 \\
min. \# episodes per show & 2 & 17 \\
max. \# episodes per show & 574 & 2665 \\
median \# episodes per show & 14.0 & 300.0 \\
avg. \# episodes per show & 43.0 & 670.9 \\
\hline
avg. \# tokens in summaries & 56.7 & 366.6 \\
avg. \# tokens in detailed recaps & 1291.7 & 2375.3 \\
avg. \# tokens in char. desc. & 702.7 & 2300.9 \\
avg. \# characters & 15.5 & 17.8 \\\hline
\end{tabular}
    \caption{Dataset statistics for \dataset}
    \label{tab:detailed_dataset_stats2}
\end{table}

We show detailed dataset statistics for \fd and \tms in Table~\ref{tab:detailed_dataset_stats2}.

\subsection{Detailed Comparison to \storium}

\begin{table}[t]
    \centering\small
    \begin{tabular}{|l|r|r|r|r|}\hline
         & uni. & bi. & tri. & four.  \\\hline
        \multicolumn{5}{|c|}{\dataset (\fandom)}\\\hline
        summ. & 34.3 & 3.4 & 0.8 & 0.3 \\
        char. desc. & 88.1 & 48.3 & 16.7 & 6.2 \\
        char. desc. $\setminus$ summ. & 54.3 & 45.4 & 16.3 & 6.1 \\
        summ. $\setminus$ char. desc. & 0.5 & 0.6 & 0.4 & 0.2 \\
        char. desc. $\cup$ summ. & 88.7 & 48.9 & 17.1 & 7.4\\\hline
        \multicolumn{5}{|c|}{\dataset (\tvmegasite)}\\\hline
        summ. & 61.7 & 14.7 & 3.0 & 1.2 \\
        char. desc. & 93.4 & 56.9 & 17.3 & 3.2 \\
        char. desc. $\setminus$ summ. & 32.7 & 44.2 & 16.1 & 3.1 \\
        summ. $\setminus$ char. desc. & 0.9 & 2.0 & 1.8 & 1.0 \\
        char. desc. $\cup$ summ. & 94.3 & 58.9 & 19.1 & 4.2 \\\hline
        \storium & 72.5 & 24.7 & 5.4 & 1.2 \\\hline
    \end{tabular}
    \caption{Fraction (\%) of n-grams in the \textit{output stories} that also appear in the source inputs. Higher fraction of overlapping n-grams indicates that the two are more directly related. For \dataset, we vary different kinds of inputs. 
    }
    \label{tab:overlap_ratio_story_generation_dataset_compare}
\end{table}

\begin{enumeratesquish}
\item Our dataset has more stories, more characters, and longer character descriptions.
\item The elements (e.g., traits of characters and story events) in \storium are constrained by cards. These cards have limited variety and are often inherited from other stories. This characteristic makes the involved characters less diverse than those in our dataset.
\item The stories in \storium often have detailed descriptions about environments and character utterances, whereas the stories in \dataset mostly narrate events that happened without these details. While this leads to shorter stories in \dataset, it also prevents the task from conflating generating events and generating other kinds of details in story generation.
\item Due to the fact that the plots in \storium are gamified and crafted by amateur writers, 89.8\% of stories in \storium are unfinished.\footnote{We label a story as unfinished if it has no completion date.} The stories in our dataset are created and refined by professional screenwriters (though the prose is written by fans, who are presumably amateurs).
\item  Stories in \storium are turn-based, where each turn is written from the perspective of a particular character and is composed by one player, so the stories often lack direct interactions among characters, unlike \dataset.
\item Unlike other story generation datasets, there is an episodic structure among the stories in \dataset, which can potentially be used to improve the modeling of characters.
\end{enumeratesquish}

\paragraph{\storium lacks direction interactions among characters.} We quantify this phenomenon in \storium by computing the frequency of occurrences of characters in each turn excluding the character that owns the turn, 
and the frequency is 0.8 on average with 50.4\% of the turns absent such occurrences.\footnote{We use string matching to detect the occurrences of characters as in the way we construct our dataset.} In contrast, TV shows advance plots by interactions among characters. 

\paragraph{Source Inputs and Output Stories are More Closely Related in \dataset than \storium.} 
To quantitatively illustrate the extent of relatedness between the source inputs and the output stories, we compute the n-gram overlap ratio (i.e., fraction of n-grams in the \textit{output stories} that also appear in the source inputs) between the inputs and outputs where higher ratios indicate that the two are more directly related. When computing the results for \storium, we use the best setting, i.e., the setting that maximizes the automatic and human evaluation scores in the original paper. We report results in Table~\ref{tab:overlap_ratio_story_generation_dataset_compare}. From the table, we see that for both \fd and \tms, using both character descriptions and summaries leads to an overlap ratio higher than \storium, suggesting that the reference stories are more reachable. Also, we observe there are more overlapping n-grams in the character descriptions than the summaries, suggesting that there is useful information that can be extracted from the character descriptions.

\subsection{Genres}

\begin{table}[t]
    \centering\small
    \begin{subfigure}[t]{0.25\textwidth}
    \begin{tabular}{|l|r|}\hline
       Genre  & Count \\\hline
Comedy  &  45  \\
Drama  &  36  \\
Action  &  32  \\
Adventure  &  27  \\
Children  &  25  \\
Fantasy  &  15  \\
Science-Fiction  &  14  \\
Crime  &  12  \\
Family  &  11  \\
Anime  &  10  \\
Romance  &  9  \\
Supernatural  &  4  \\
Horror  &  3  \\
Thriller  &  3  \\
Mystery  &  3  \\
Music  &  2  \\
Medical  &  2  \\
Legal  &  1  \\
History  &  1  \\\hline
    \end{tabular}
    \end{subfigure}%
    \begin{subfigure}[t]{0.25\textwidth}
    \begin{tabular}{|l|r|}\hline
       Genre  & Count \\\hline
    Drama  &  9  \\
Romance  &  5  \\
Family  &  3  \\
Medical  &  1  \\\hline
    \end{tabular}
    \end{subfigure}
    \caption{Genres for TV shows in \fandom (left) and \tvmegasite (right).}
    \label{tab:genre}
\end{table}

\begin{table}[t]
    \centering\small
    \begin{tabular}{|l|r|}\hline
Genre & Count (Fraction) \\\hline
Action & 5385 (15.0\%) \\
Comedy & 5214 (14.5\%) \\
Drama & 4793 (13.4\%) \\
Adventure & 4220 (11.8\%) \\
Children & 3280 (9.1\%) \\
Science-Fiction & 2524 (7.0\%) \\
Anime & 2239 (6.2\%) \\
Fantasy & 1949 (5.4\%) \\
Romance & 1408 (3.9\%) \\
Family & 1355 (3.8\%) \\
Crime & 1139 (3.2\%) \\
Supernatural & 729 (2.0\%) \\
Medical & 480 (1.3\%) \\
Horror & 309 (0.9\%) \\
Mystery & 299 (0.8\%) \\
Thriller & 240 (0.7\%) \\
Music & 220 (0.6\%) \\
History & 47 (0.1\%) \\
Legal & 19 (0.1\%) \\\hline
    \end{tabular}
    \caption{Genres in \fandom and their corresponding numbers and percentages of episodes.}
    \label{tab:fandom_genre_freq}
\end{table}

\begin{table}[t]
    \centering\small
    \begin{tabular}{|l|r|}\hline
Genre & Count (Fraction) \\\hline
Drama & 15430 (44.6\%) \\
Romance & 8341 (24.1\%) \\
Family & 7686 (22.2\%) \\
Medical & 3144 (9.1\%) \\\hline
    \end{tabular}
    \caption{Genres in \tvmegasite and their corresponding numbers and percentages of episodes.}
    \label{tab:tvmegasite_genre_freq}
\end{table}

To demonstrate the diversity of topics covered in \dataset, we report distributions of genres in Table~\ref{tab:genre}, Table~\ref{tab:fandom_genre_freq}, and Table~\ref{tab:tvmegasite_genre_freq}.\footnote{The category information is from TVMaze (\url{https://www.tvmaze.com/}) where a show may correspond to multiple genres.} While \tvmegasite has a relatively small number of genres, \fandom covers far more genres with the most frequent occupying only 15\% of episodes.

\section{Decoding}
For both story generation and summarization, we use a batch size of 200, beam search of size 5 with n-gram blocking \cite{paulus2018a} where probabilities of repeated trigrams are set to 0 during beam search. We did not find nucleus sampling \cite{Holtzman2020The} leading to better generation quality (i.e., fluency and faithfulness to the summaries and the recaps) than beam search with n-gram blocking, possibly due to the fact that our models generate at the sentence level.

\subsection{Hyperparamters}
Because our plot-to-text models work at the sentence level, leading to many training instances for both \fd and \tms (i.e., 0.5 million and 1.5 million sentences respectively), we train these plot-to-text models for 10 epochs without early stopping. During generation, we set the minimum number of decoding steps to 24 and the maximum number of decoding steps to 48.

As for summarization, we benchmark pretrained BART-base, BART-large \cite{lewis-etal-2020-bart}, and Pegasus \cite{pegasus-zhang20ae}. As the average length of the detailed recaps is much longer than the default maximum sequence length of these pretrained models, we extend the maximum sequence length to 4096. When doing so, we randomly initialize new positional embeddings for BART. Since Pegasus uses Sinusoidal positional embeddings, we simply change the default value of maximum sequence length. We train the models for 15 epochs and perform early stopping on the dev set perplexities. During generation, we limit the minimum decoding step to be 50 and 300, and the maximum decoding step to be 100 and 600 for \fd and \tms respectively. The minimum decoding steps roughly match the average length of the summaries in \fd and \tms.

\subsection{Computing Resources and Number of Parameters}
We train our models on a RTX 2080Ti, which takes approximately 8 days to complete training the story generation model that uses plots and character descriptions and summaries as input. It takes approximately 12 hours to train the reverse model.

As for numbers of parameters for each models, since we do not modify model architectures, their numbers of parameters should be similar to those of the pretrained checkpoints we used.

\subsection{Summarization Dataset}
By considering generating the brief summary from the detailed episode recap, we create an abstractive summarization dataset \datasetsum. In this dataset, we simply use the detailed episode recap as the source input and the summary as the target output and leave the integration of character descriptions to future work. In experiments, we use the abstractive summarization models trained on this dataset for evaluating the faithfulness of generated stories.

We characterize \datasetsum by comparing it to other abstractive summarization datasets, finding that it favors abstractive approaches. In addition, unlike most other summarization datasets, our dataset focuses on stories. These two characteristics make our dataset a potentially valuable contribution for the summarization community.

\subsection{Comparison to Other Summarization Datasets}
\label{sec:compare_to_other_summ_datasets}

\begin{table}[t]
    \centering\small
    \setlength{\tabcolsep}{5pt}
    \begin{tabular}{|l|r|r|r|r|r|r|}\hline
         & uni. & bi. & tri. & four. & src. & tgt. \\\hline
        \multicolumn{7}{|c|}{\datasetsum}\\\hline
        \fd & 73.0 & 26.7 & 8.1 & 3.0 & 1.3k & 56.7 \\
        \tms & 85.0 & 40.1 & 13.6 & 5.8 & 2.4k & 366.6 \\\hline
        \multicolumn{7}{|c|}{Other summarization datasets}\\\hline
        XSum\textsuperscript{\dag} & 64.2 & 16.6 & 4.5 & 1.5 & 431.1 & 23.3 \\
        CNNDM\textsuperscript{\S} & 80.5 & 43.1 & 25.6 & 17.2 & 810.6 & 56.2\\
        MNews\textsuperscript{\S} & 82.2 & 42.9 & 24.3 & 17.7 & 2.1k & 264.7 \\\hline
    \end{tabular}
    \caption{Fraction (\%) of n-grams in the \textit{output summaries} that also appear in the inputs, and the average numbers of tokens for the inputs and outputs. Datasets with smaller fractions of overlapping n-grams tend to favor abstractive summarization approaches. Results marked by \dag\xspace and \S\xspace are from \citet{narayan-etal-2018-dont} and \citet{fabbri-etal-2019-multi} respectively.}
    \label{tab:overlap_ratio_summarization_dataset_compare}
\end{table}

\begin{table*}[t]
    \centering\small
    \begin{tabular}{|l|r|r|r|c|}\hline
         & \# instances & \# tokens (input) & \# tokens (summary) & Domain \\\hline
        \multicolumn{5}{|c|}{Long-form text summarization datasets}\\\hline
        Multi-News \citep{fabbri-etal-2019-multi} & 56.2k & 2103.5 & 264.7  & News \\
        RottenTomatoes \citep{wang-ling-2016-neural}  & 3.7k & 2124.7 & 22.2 & Reviews \\
        arXiv \citep{cohan-etal-2018-discourse}  & 215k & 4938.0 & 220.0 & Science \\
        PubMed \citep{cohan-etal-2018-discourse} & 113k & 3016.0 & 203.0 & Science \\
        GovReport \citep{huang-etal-2021-efficient} & 19.5k & 9409.4 & 553.4 & Government Reports \\
        \datasetsum & 29.0k & 1868.7 & 221.6 & Television Series \\
        \hline
        \multicolumn{5}{|c|}{Dialogue-related summarization datasets}\\\hline
        SAMSum \citep{gliwa-etal-2019-samsum} & 16.4k & 83.9 & 20.3 & Chitchat \\
        QMSum \citep{zhong2021qmsum} & 1.8k & 9069.8 & 69.6 & Meetings \\
        MediaSum \citep{zhu2021mediasum} & 463.6k & 1553.7 & 14.4 & News Interviews \\
        ForumSum \citep{khalman-etal-2021-forumsum-multi} & 4.1k & 303.5 & 36.0 & Forum Messages \\
        SummScreen \citep{chen2021summscreen} & 26.9k & 6612.5 & 337.4 & Television Series  \\\hline
    \end{tabular}
    \caption{Statistics for datasets focusing on abstractive summarization for long-form text or dialogue. The numbers are averaged over instances.
    }
    \label{tab:other_summarization_dataset_compare}
\end{table*}

We briefly compare \datasetsum to  three summarization datasets: CNNDM \cite{NIPS2015_afdec700}, XSum \cite{narayan-etal-2018-dont}, and MNews \cite{fabbri-etal-2019-multi}. For \datasetsum, we simply use the detailed episode recap as the source input and the summary as the target output and leave the integration of character descriptions to future work. We report n-gram overlap ratio (i.e., fraction of n-grams in the \textit{output stories} that also appear in the source inputs) and length statistics in Table~\ref{tab:overlap_ratio_summarization_dataset_compare}. The n-gram overlap ratio is usually used as an indicator of the abstractiveness of a summarization dataset. Lower ratio indicates a higher degree of abstraction. CNNDM favors extractive approaches, whereas XSum is known for it is abstractiveness. We also compare to MNews because it shares similar input and output lengths as our dataset. As shown in the table, our dataset tends to be more abstractive. In addition, unlike other summarization datasets, our dataset focuses on stories. These two characteristics make our dataset a potentially valuable contribution for the summarization community. Comparison to other abstractive summarization datasets is in  Table~\ref{tab:other_summarization_dataset_compare}.

\subsection{Summarization Results}

\begin{table}[t]
    \centering\small\setlength{\tabcolsep}{6pt}
    \begin{tabular}{|l|r|r|r|r|}\hline
         & BL & R1 & R2 & R3  \\\hline
         \multicolumn{5}{|c|}{\fandom} \\\hline
        Extractive oracle & 8.3 & 37.0 & 11.3 & 30.9 \\
        BART-base & 5.2 & 31.2 & 7.3 & 25.5 \\
        BART-large & 5.4 & 30.7 & 7.6 & 25.3 \\
        Pegasus & \bf 5.7 & \bf 31.3 & \bf 7.7 & \bf 25.6 \\\hline
         \multicolumn{5}{|c|}{\tvmegasite} \\\hline
        Extractive oracle & 16.9 & 55.8 & 20.9 & 53.6 \\
        BART-base & \bf 8.3 & \bf 43.8 & \bf 12.6 & \bf 42.3 \\
        BART-large & 8.1 & 43.2 & 12.3 & 41.8 \\
        Pegasus & 7.7 & 43.5 & \bf 12.6 & 42.1 \\\hline
    \end{tabular}
    \caption{\datasetsum test results for summarizing detailed episode recaps. The best result in each column for each domain (excluding the oracle) is boldfaced.}
    \vspace{-1.5em}
    \label{tab:summarization_main_result}
\end{table}

We report BLEU and ROUGE scores for summarization in Table~\ref{tab:summarization_main_result}. We report the performance of an extractive oracle where for each sentence in the reference summary, we pick a sentence in the detailed episode recap that maximizes the average of the three ROUGE scores compared against the summary sentence. While recent pretrained models, such as Pegasus, have outperformed the oracle extractive approaches by a large margin on datasets with a high degree of abstractiveness (e.g., XSum \cite{narayan-etal-2018-dont}), the results in the table show that our dataset is still challenging for these pretrained models.

It is also interesting to see that while Pegasus is best for \fandom, it is not best on \tvmegasite. This may be because \tvmegasite has longer summaries than \fandom. Also, the performance of BART-base is comparable to that of BART-large on \fandom and is better than Pegasus and BART-large on \tvmegasite. This is likely because there is a limited amount of data with similar writing style in pretraining, resulting in little benefit of using larger models for this downstream task. We provide this abstractive summarization task to the community as a challenging dataset for future work.

\subsection{Details of Semantic Role Labeling }

We use a pretrained model from \citet{shi2019simple} to generate SRL tags of the detailed episode recaps. We eliminate the SRL tags for sentences that do not contain $\langle\text{ARG0}\rangle$ or only contain pronouns to avoid ambiguity. For each sentence, we also only keep the SRL tags that correspond to the first verb that appears in the sentence to avoid the SRL tags being too specific, so there will be a balanced burden between the text-to-plot model and the plot-to-text model. In addition, following \citet{goldfarb-tarrant-etal-2020-content}, we discard SRL tags of generic verbs.

The list of verbs we discard is as follows: ``is'', ``was'', ``were'', ``are'', ``be'', ``'s'', ``'re'', ``'ll'', ``can'', ``could'', ``must'', ``may'', ``have to'', ``has to'', ``had to'', ``will'', ``would'', ``has'', ``have'', ``had'', ``do'', ``does'', ``did''.

We also eliminate arguments that are longer than 5 tokens.

\subsection{Quantifying Negations in Generated Stories.} To quantify the observation that the models tend to generate sentences that use the word ``but'' to negate what has been said in the earlier part of the sentence, we compute the frequency of the word ``but'' per sentence for the reference stories, ``oracle plot+summary'', and ``oracle plot+summary+oracle char.~desc.''. The results are  0.13, 0.53, and 0.24, respectively.

\subsection{Human Evaluation Instructions}

\begin{table}[t]
    \centering\small
    \begin{tabular}{|l|p{0.3\textwidth}|}\hline
       Task Description  & We are going to ask you two questions to evaluate the relevancy and interestingness of beginnings of English stories given a brief summary.  \\\hline
       Question 1  & Please choose which story you think is more relevant to the summary. By relevant, we mean specifically: Do they share similar topic, details or character names?  \\\hline
       Question 2 & Please choose which story you think is more interesting. By interesting, we mean specifically: Does the story advance the plot smoothly instead of dwelling on each event separately? Does the story have interesting and reasonable details? Please disregard issues of grammaticality and fluency of the presented text. \\\hline
    \end{tabular}
    \caption{Instructions to crowd-workers on Amazon Mechanical Turk.}
    \label{tab:amt_instruct}
\end{table}

In Table~\ref{tab:amt_instruct}, we show the instructions presented to the annotators. In particular, we show the annotators beginnings of two stories and a summary and ask them to choose which story is better for the two questions in Table~\ref{tab:amt_instruct}.

\subsection{Generation Examples}

\begin{table*}[t]
    \centering
    \small
\begin{tabular}{|p{0.3\textwidth}|p{0.3\textwidth}|p{0.3\textwidth}|}\hline
\multicolumn{1}{|c|}{Summary} & \multicolumn{1}{|c|}{Reference} & \multicolumn{1}{|c|}{Oracle plot + summ. + oracle char. desc.} \\\hline
Elfman and Evergreen , after much struggle , lose to Rustyrose and his imagination - based Magic . In the meantime , Gray , Lucy , Cana and Loke have been overpowered by Caprico alone . Loke decides to take him on by himself because of his mysterious Magic . 
& Elfman and Evergreen encounter a cliff as they run away from Rustyrose 's Belcusas the Thunderclap . Rustyrose appears shortly afterwards and expands on the idea of " The Ultimate World of Magic " , saying that all those who can not use Magic and the trash in the guilds are useless . However , Rustyrose outsmarts them despite their teamwork . He proceeds to explain the nature of his Magic , Arc of Embodiment , and then finishes the two off with Tower of Dingir . Meanwhile , Natsu awakes with his scarf and clothes reverted to normal by Wendy . Upon thanking Wendy , he picks up the scent of Zalty , the masked man he met back in Galuna Island . On the Magic Council warship , Doranbolt , formerly known as Mest Gryder , appears before Lahar who is ordering people to repair the damaged ship . Doranbolt reports to Lahar about the events on Tenrou Island Lahar warns him that Gran Doma , the Chairman , is very strict and would likely eliminate those who he believes are wrongdoers without a trial and he might resort to using Etherion once more . In the battle against Caprico , Loke tells the others to leave and look for Caprico 's teammates while he battles Caprico himself . At first , Lucy did n't want to leave . This action makes Caprico slightly recognize her . He discovers that Caprico 's Magic enables him to capture and summon humans and that Caprico has no master .
& Elfman and Evergreen encounter a cliff as they run away from the danger of the Rustyrose 's attacks . Evergreen explains that " all those who use Magic " will use the power of the Magic of the Seven . However , Rustyrose outsmarts them , stating that their Magic is useless against them , as they have already been defeated . He then finishes the two with Tower of Dingling , then uses his Magic on his left arm , freeing Freed . Meanwhile , Natsu and Gray , both reverting back to normal by Wendy 's scarf , prepare to leave for Fairy Tail . After thanking Wendy , he realizes that he is the one who set up the destruction of the island , as he does not want to be a part of his family 's legacy . Meanwhile , Lahar appears from the sky , ordering people to repair the damaged ship , as the Celestial Spirit Mage appears before them . Afterwards , Doranbolt reports to Lahar about the destruction of the Heart Tree from the Onyrium Cascade . However , Lahar warns him that the Celestial Spirit Mage , who is known as Elfman , would have the power to kill him . and he is once more able to resort to using Etherion once more , once more using his Magic , but once again , Etherion is more powerful than ever . Loke then tells the others to leave Gray 's body in the Celestial Spirit Graveyard while he deals with Evergreen . At first , Lucy did n't want to leave , but then she did , and they started to have a friendly relationship . This action makes Capricorn slightly recognize her and makes the Magic Mage lose control of his Magic . . . which makes her lose her Magic . Capricorn 's Magic enables him and humans to capture and summon humans , stating that he has no choice .
\\\hline

\end{tabular}
\caption{Complete generation for the first examples.}
\label{tab:complete_example_first}
\end{table*}

\begin{table*}[t]
    \centering
    \small
\begin{tabular}{|p{0.3\textwidth}|p{0.3\textwidth}|p{0.3\textwidth}|}\hline
\multicolumn{1}{|c|}{Summary} & \multicolumn{1}{|c|}{Reference} & \multicolumn{1}{|c|}{Oracle plot + summ. + oracle char. desc.} \\\hline
Chuck is preparing for his new club opening and enlists Serena 's help , but Blair begins to feel left out . Jenny , the new Queen at Constance , struggles between proving herself and her friendship with Eric , and Dan feels inferior after watching one of Olivia 's movies . Meanwhile , Lily tries to respect Rufus ' Halloween traditions .
& Lily wants to get rid of Jenny 's old clothes , including one of the dresses she made , but Jenny insists that all of her clothes fit her perfectly and wants to keep them . At The Empire , Chuck tells Blair how the hotel is doing in terms of bookings . Blair says that if he opens the club , it will bring in business and make bookings go up . She suggests an 80s themed party but Chuck shoots down the idea . Blair , suspicious that Chuck is still angry over her lying to him to get the NYU freshman toast ( in Enough About Eve ) , tells him that she apologized and they should move on . He explains that he needs to do things his way and leaves to meet with his accountant . When he 's gone , Blair calls a party planner to plan the opening of the club . At the loft , Nate has brought Dan the Endless Knights trilogy starring Olivia . He tells Dan about how Patrick Roberts , who plays her boyfriend in the movies , was actually her boyfriend in real life . Dan seems unfazed , until he gets a Gossip Girl blast with a picture of Olivia getting some condoms . She 's annoyed because Serena did n't break up Olivia and Dan like she was asked to . KC explains that Olivia needs to be with Patrick , because otherwise he 's headed for a serious career stall and becoming irrelevant . She instructs Serena again to break the two up , or else she 'll be stuck running pointless errands for the rest of her life . On the way to school , Eric and Jenny discuss Halloween . Jenny tells him they have to find a party or else risk handing out candy with Rufus , when they run into her minions . Eric goes off to find Jonathan and Jenny begins bossing her minions around . She tells them to move , not wanting to risk looking weak to her minions , and they do . As they walk off , Jonathan tells Eric that she 's changing . Eric defends Jenny , saying it 's only a mask for school and Jonathan Back at the loft , Nate and Dan are watching the Endless Knights trilogy . Dan watches one of the sex scenes and begins to get uncomfortable . Nate asks if he 's okay , and Dan tries to play it cool when Olivia calls . He pretends to be sick to buy some time to finish the films and think things over . Blair goes to see Serena at KCs office . She tells her that she already hired a party planner , but Serena is confused since Chuck wanted to wait until after the holidays to open the club ... &
Jenny wants to keep them , and all of her friends get out of the way to get her out of there . At The Empire , Chuck tells Blair how Serena is coming back to town in Los Angeles to see her ex . Blair says that if he opens the club , it 's an opening opportunity , and Chuck says he will . She suggests an 80s themed party , but Chuck is not interested , as she 's taking a break from the party . Afterwards , she tells him that she knows about her lying to him over the phone and that she is n't going to give up . He tells her that he needs to do things his way , and she agrees to do it the next day before he leaves . Back at the VDW 's , Blair calls a party planner to tell everyone that she 's going to the party . At the loft , Nate has brought Dan everything he 's ever wanted to know about Rufus and Jenny . He tells Dan how Serena broke up with him at the bar , and that she 's still in love with him . Dan replies that he gets a Gossip Girl blast every day , and that 's why she 's so obsessed with The Spectator . Serena threatens to break up Olivia and Dan if she does n't break up with him at some point . At the VDW 's , Olivia needs to be with Patrick because she feels like he needs her for something . She instructs Serena to break the two up , and once she does , the two begin to break up . On the way back to school , Eric and Jenny discuss Halloween and agree that it 's the perfect way to celebrate . Jenny tells him they have to figure out a way to get through the party , and Rufus asks where they 're going . Eric goes off to find Jonathan , and he 's surprised to find her at the table with all of her friends . She tells them to move , and they do , and Eric tells her not to do anything stupid and to do what 's right . At The Palace , Jonathan tells Eric that she 's changing , and she tells him that they should move in together . Eric defends Jenny , saying it 's only a matter of time before he 's exiled , and Serena says it is . Back at the loft , Nate and Dan are watching the Endless Knightsnorketomy and discuss Dan 's plans for the night . Dan watches one of the sex scenes and begins to get jealous of Chuck 's lack of sex appeal , but Dan breaks up with him and leaves . Nate calls and asks if he 's okay , and Rufus says it 's fine if he leaves . He pretends to be sick and she goes over to ask if he needs to spend the time with her and Rufus ...
\\\hline

\end{tabular}
\caption{Part of the generated text for the second examples. Due to space constraints, we only show the beginnings of the generated stories and the reference story.}
\label{tab:complete_example_second}
\end{table*}
We show generation examples in Table~\ref{tab:complete_example_first} and Table~\ref{tab:complete_example_second}.

\end{document}